# Adjustment Criteria in Causal Diagrams: An Algorithmic Perspective

**Johannes Textor**
Institute for Theoretical Computer Science
University of Lübeck, Germany
textor@tcs.uni-luebeck.de

**Maciej Liśkiewicz**
Institute for Theoretical Computer Science
University of Lübeck, Germany
liskiewi@tcs.uni-luebeck.de

## Abstract

Identifying and controlling bias is a key problem in empirical sciences. Causal diagram theory provides graphical criteria for deciding whether and how causal effects can be identified from observed (nonexperimental) data by covariate adjustment. Here we prove equivalences between existing as well as new criteria for adjustment and we provide a new simplified but still equivalent notion of $d$-separation. These lead to efficient algorithms for two important tasks in causal diagram analysis: (1) listing minimal covariate adjustments (with polynomial delay); and (2) identifying the subdiagram involved in biasing paths (in linear time). Our results improve upon existing exponential-time solutions for these problems, enabling users to assess the effects of covariate adjustment on diagrams with tens to hundreds of variables interactively in real time.

## 1 Introduction and Motivation

A notorious problem affecting probabilistic reasoning in causal structures is bias. For instance, we might study whether regular coffee drinking (C) increases the risk of suffering a heart attack (H). Such a study might be compromised by an unobserved genetic predisposition (U) that causes an increased preference for coffee drinking but also for smoking (S), which does increase the risk to suffer a heart attack. These causal influences can be modeled as a *causal diagram* $G = C \leftarrow U \rightarrow S \rightarrow H$, whose arrows indicate the directionality of causal relationships between the variables of interest [9]. The paths from U to both C and H in this diagram indicate that an observed relation between C and H will be *confounded* by U, which may obscure or artificially increase the putative causal effect. This bias can be controlled by *adjustment* for S, e.g. by averaging separate effect estimates for smokers and non-smokers. We can represent adjustment in the diagram by labeling S: C ← U → ⬛S⬛ → H. This labeling *blocks the biasing path* from C to H.

To avoid bias, it may seem advisable to adjust for all covariates in our study. Unfortunately, adjustment can also *create* bias. A folklore example is the following: If we ask Harvard students for their grades and their parents' income, we may well find an *inverse* correlation between the two, which could lead us to the interesting hypothesis that rich people have dumber than average children. However, a more likely explanation is that having rich parents (R) or being smart (S) both increase the odds of getting to Harvard (H). This hypothesis corresponds to the diagram R → ⬛H⬛ ← S, in which H is labeled because we implicitly adjusted for H by restricting our study to Harvard students only. Because H is a common descendant of R and S, this *opens a biasing path* between R and S, which leads to so-called *Berksonian bias* [2] – hence, our "interesting" observation is merely an artifact. A further example where confounding and Berksonian bias combine to so-called *M-bias* is shown in Figure 1. In general, any type of bias that can be expressed in the formal framework of causal diagrams corresponds to a certain type of path, called a *biasing path* [10]. Within the mathematical theory behind causal diagrams [9], graphical identification criteria have been derived that tell us whether – and how – we may dissect the causal effect from bias by covariate adjustment, provided that we know the causal relationships between exposure, outcome, and covariates in our study. Sets of covariates that allow identifying the causal effect are called *adjustments*. In this paper, we are concerned with *minimal* adjustments, which do not contain other adjustments as proper subsets (e.g., {FI} and {MD, MR} but not {FI, MD, MR} in Figure 1). Minimal adjustments are important because adjusting for too many variables may decrease statistical power.

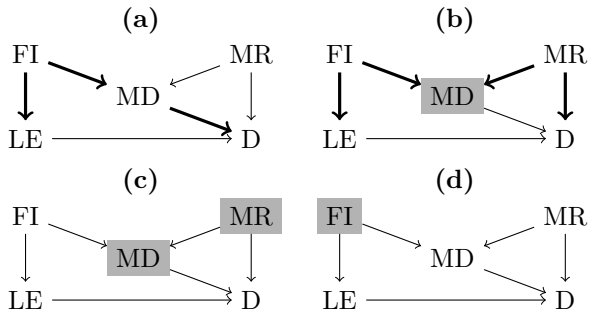

Figure 1: Causal diagram [10, Chapter 12] for a study of the effect of low education (LE) on diabetes risk (D) with the covariates family income (FI), mother's genetic risk to develop diabetes (MR), and mother's diabetes (MD). The unadjusted estimate (a) is biased due to the common ancestor FI – bias "flows" via the biasing path LE ← FI → MD → D (bold edges). Adjustment for MD (b) blocks this biasing path, but opens a new one because FI and MR become correlated. The minimal adjustments {MD, MR} (c) and {FI} (d) close all biasing paths.

## 2 Contributions

If done by hand, testing the graphical criteria for adjustment is a cumbersome and error-prone process unless the diagram is very small. This is because most criteria are stated as path properties, and even diagrams with merely tens of variables and edges can contain millions of paths[1]. The analysis of such diagrams lends itself well to automation by a computer program. In 2010, at least three programs were presented for analyzing causal diagrams: The *DAG program* by Knüppel [6]; *dagR* by Breitling [3]; and *Commentator* by Kyono [7]. However, even these programs still have their problems with large diagrams [3], to the extent that diagrams with tens of variables can be intractable.

We argue that the root of these problems is that existing criteria do not lend themselves well to algorithmic implementation. For example, the aforementioned programs contain functions to list all minimal adjustments. They do this by generating all possible covariate sets, and then testing for each set whether Pearl's *back door criterion* (to be defined later) is satisfied. For a diagram with 30 variables, this means that $2^{30}$ covariate sets may have to be tested – a very large number even for a computer program. Here, we will develop criteria that lead naturally to more efficient algorithms for solving such problems.

---

[1]The author knows of an incident where a student was asked to list all paths for a diagram with 10 variables and 37 edges. It took three months.

After defining the preliminaries in Section 3, the presentation of our results is organized in two parts: the analysis of adjustment and *d*-separation criteria in Section 4, and the algorithmic applications in Section 5. Thus causality theorists who may be interested in the discussion of the criteria, but not in the algorithms, can focus on to Section 4. Specifically, our contributions address the two following problems:

*(1) Enumeration of all minimal adjustment sets.* The first complete (i.e., necessary and sufficient) criterion for validity of covariate adjustment was recently given by Shpitser et al. [12]. We prove that when restricting our attention to *minimal* adjustments, this is equivalent to Pearl's sufficient, but not necessary, *back-door criterion* [9]. This way we also obtain a complete criterion for minimal covariate adjustment in terms of the *moral graph* [8]. Our results hold for acyclic causal diagrams (DAGs) and extend to the case where exposure and outcome are sets of variables, with the minor restriction that the diagrams be loop-free with respect to the exposure (i.e. there exists no causal path that starts and ends in $X$).

Because the moral graph is an undirected graph, this leaves us with a minor variation of a standard graph problem: finding vertex separators of an undirected graph. In Section 5.1, we present an algorithm that outputs the set of all minimal adjustments (which may have exponential size) using only polynomial time per element output. This improves upon existing algorithms that either require exponential worst-case time before producing any output (such as those used in the aforementioned programs), or terminate in polynomial time but output only one solution (e.g. [1, 15]).

*(2) Identifying all biasing paths.* It is often desirable to determine via which covariates a bias is actually mediated (bold paths in Figure 1a,b). In large diagrams, this can be difficult as the somewhat intricate definition of biasing paths allows sometimes, but not always, to use edges in the opposite direction. In Section 4.2, we show that one can transform (by stripping the heads from certain arrows) a diagram $G$ adjusted for $Z$ into a mixed graph $G^Z_\phi$ with the property that open paths in $G$ bijectively correspond to *forks* in $G_\phi$; a fork is a pair of directed paths connected by an undirected path. This characterization yields an algorithm that finds all biasing paths (i.e., labels all edges lying on biasing paths like in Figure 1a,b) in linear time (Section 5.2). To our knowledge, no previous algorithm exists that addresses this problem; to facilitate bias diagnosis in causal diagrams, the *DAG program* and *dagR* instead resort to listing all biasing paths – which quickly becomes intractable in even modestly sized diagrams, as discussed above.

## 3 Preliminaries

We use the following terms from the causal diagram literature, most of which are identical to their standard graph theory counterparts. However, we would like to point out to readers unfamiliar with causal diagrams that the notion of a *path* in a causal diagram is in fact a bit different from the usual notion of a path in a directed graph (termed here a *directed* or *causal path*).

**Graphs and variables.** A *digraph* is a tuple $G = (V, E)$ of vertices (nodes) $V$ and directed edges $E \subseteq \{(u,v) \mid u, v \in V, u \neq v\}$. More generally, a *mixed graph* is a tuple $G = (V, E)$ of vertices $V$ and directed or undirected edges $E \subseteq \{(u,v) \mid u, v \in V, u \neq v\} \cup \{\{u,v\} \mid u, v \in V, u \neq v\}$. Because the vertices in causal diagrams represent observed variables, we use the terms "vertex", "node" and "variable" interchangeably. Given two vertex sets $X$ and $Y$, the other vertices from $V \setminus (X \cup Y)$ are also called *covariates*. We will often specify a subset of *latent covariates* $L \subseteq V \setminus (X \cup Y)$, for which we cannot adjust.

**Paths.** A *path* of length $k-1$ is a sequence of vertices $v_1, \ldots, v_k$ in which each vertex occurs at most once, and for all $i \in \{1, \ldots, k-1\}$, $v_i$ and $v_{i+1}$ are connected by a directed edge $(v_i, v_{i+1})$ or $(v_{i+1}, v_i)$ or an undirected edge $\{v_i, v_{i+1}\}$. A path can have length 0. A path from $x \in V$ to $y \in V$ is called *causal* or *directed* if it only contains directed edges pointing away from $x$, and is called *biasing* otherwise. A directed or mixed graph is called *acyclic* if there is no directed path of nonzero length from a vertex to itself; "directed acyclic graph" is abbreviated by "DAG".

**Descendants and Ancestors.** If there is a directed path from $x$ to $y$ then $x$ is an *ancestor* of $y$ and $y$ is a *descendant* of $x$. The ancestor set $An(X)$ of a vertex set $X$ contains all ancestors of vertices in $X$ (this includes $X$). Analogously, the descendant set $De(X)$ is the set of all descendants of any node in $X$. Given a graph $G = (V, E)$ and a vertex set $W \subseteq V$, the *ancestor graph* $G[An(W)]$ is the subgraph of $G$ consisting only of the vertices in $An(W)$ and all edges between them.

In addition to these standard concepts, we use the following notions from the causal diagram literature.

**Definition 3.1** (*d*-Connectivity and *d*-Separation [9]). A path $\pi = v_1, \ldots, v_k$ in a DAG $G = (V, E)$ is called *d-connected, active* or *open* with respect to $Z \subseteq V$ if (1) for all subsequences of $\pi$ of the form $v_{i-1} \leftarrow v_i \rightarrow v_{i+1}$, $v_{i-1} \rightarrow v_i \rightarrow v_{i+1}$ or $v_{i-1} \leftarrow v_i \leftarrow v_{i+1}$, the middle vertex $v_i$ is not in $Z$; (2) for all subsequences of $\pi$ of the form $v_{i-1} \rightarrow v_i \leftarrow v_{i+1}$, the middle vertex $v_i$ is in $An(Z)$. If one of these conditions does not hold then the path is called *blocked* by $Z$. Two disjoint vertex sets $X, Y \subseteq V$ are *d-connected* with respect to $Z$ if there exists a *d*-connected path from some $x \in X$ to some $y \in Y$. If such a path does not exist then $X$ and $Y$ are said to be *d-separated* by $Z$.

**Definition 3.2** (Moral Graph [8]). Given a directed graph $G$, the *moral graph* $G^m$ is the undirected graph obtained by transforming $G$ as follows: (1) For all pairs of edges of the form $(u, v), (w, v)$, add an undirected edge $u, w$ to $G$. (2) Substitute every directed edge $(u, v)$ by an undirected edge $\{u, v\}$.

**Definition 3.3** (Back-Door Graph). Given a directed graph $G = (V, E)$ and a vertex set $X \subseteq V$, the *back-door graph*, denoted as $G_{\underline{X}}$, is obtained by removing all edges $(u, v) \in E$ where $u \in X$ and $v \notin X$.

To define the causal effect of $X$ on $Y$ we use Pearl's $do(x)$ notation [9], which intuitively corresponds to an idealized experiment in which the variables in $X$ can be set to given values; in the causal diagram, this would correspond to removing all edges entering $X$, disconnecting directed influences stemming from the parent variables of $X$. We will denote the resulting graph as $G_{\bar{X}}$.

**Definition 3.4** (Adjustment [9]). Given a DAG $G = (V, E)$ and three pairwise disjoint vertex sets $X, Y, Z \subseteq V$, the set $Z$ is called *covariate adjustment for estimating the causal effect from $X$ to $Y$*, or simply *adjustment*, if $P(y \mid do(x)) = \sum_z P(y \mid x, z) P(z)$ in every model that induces $G$. $Z$ is a *minimal adjustment* if no proper subset of $Z$ is also an adjustment.

## 4 Criteria for Adjustment and Biasing Paths

This section is organized in two parts. In the first part, we show that the sound but incomplete *back-door criterion* for identifiability of the causal effect becomes equivalent to the slightly more complex, but sound and complete criterion recently proposed by Shpitser et al. [12] if we consider only *minimal* adjustments. These results hold for all directed acyclic graphs with the following property:

**Definition 4.1** (*X*-Loop-Freeness). A DAG $G = (V, E)$ is *X-loop-free* with respect to the vertex set $X \subseteq V$ if it contains no directed path $\pi = v_1, \ldots, v_k$, with $k \geq 3$, $v_1, v_k \in X$ and $v_2, \ldots, v_{k-1} \notin X$.

$G$ is always $X$-loop-free if $X$ is a singleton set. For an example DAG $G$ that is not $X$-loop-free see Figure 2.

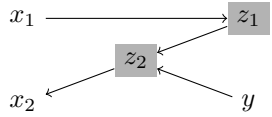

Figure 2: An example of a causal diagram $G$ that is not $X$-loop-free with respect to the vertex set $X = \{x_1, x_2\}$. For $G$ and the vertex set $Y = \{y\}$ the set $Z = \{z_1, z_2\}$ fulfills the adjustment criterion by Shpitser et al. [12] but there is no set $Z'$ fulfilling the back-door criterion by Pearl [9]. Moreover, no $Z'$ satisfies the moral graph criterion (Definition 4.3 below) as well.

### 4.1 Equivalence of Adjustment Criteria in the $X$-Loop-Free, Minimal Case

The following criterion was recently given by Shpitser et al. [12], and is the first complete criterion for covariate adjustment.

**Theorem 4.2** (Adjustment Criterion [12]). *Given a causal diagram $G = (V, E)$ and three pairwise disjoint vertex sets $X, Y, Z \subseteq V$ the following two statements are equivalent:*

1. *$Z$ is an adjustment in $G$ with respect to $X$ and $Y$.*
2. *(Adjustment Criterion) (i) No element in $Z$ is a descendant in $G_{\bar{X}}$ of any $w \in V \setminus X$ on a causal $X$-$Y$-path[2] (forbidden vertex), and (ii) all non-causal paths in $G$ from $X$ to $Y$ are blocked by $Z$.*

The following two criteria from the literature are sound, but not complete.

**Definition 4.3.** Let $G = (V, E)$ be a given causal diagram and let $X, Y, Z \subseteq V$ be disjoint sets of nodes.

- (Back-door Criterion [9]) $Z$ satisfies the back-door criterion relative to $(X, Y)$ if (i) no element in $Z$ is a descendant of $X$ and (ii) $Z$ d-separates $X$ and $Y$ in $G_{\underline{X}}$.
- (Moral Graph Criterion[3]) $Z$ satisfies the moral graph criterion relative to $(X, Y)$ if (i) $Z \subseteq An(X \cup Y) \setminus De(X)$ and (ii) $Z$ separates $X$ and $Y$ in the ancestor moral graph $(G_{\underline{X}}[An(X \cup Y)])^m$.

It is easy to see that the criteria of Theorem 4.2 are not equivalent with the back-door criterion. For example the graph $z \leftarrow x \rightarrow y$ does not fulfill the back-door criterion but both criteria of Theorem 4.2 are satisfied. However, if we restrict our attention to *minimal* vertex sets $Z$ then all criteria became equivalent if we assume additionally that a DAG $G$ is $X$-loop-free. As usually we say here that $Z$ is a minimal set satisfying a property $\mathcal{P}$ if $Z$ satisfies $\mathcal{P}$ and no proper subset $Z' \subsetneq Z$ satisfies $\mathcal{P}$.

---
[2]Shpitser et al. [12] write instead "... on a proper causal $X$-$Y$-path". However, in the $X$-loop-free DAGs we consider, all causal $X$-$Y$-paths are proper.

[3]This criterion is an extension of the work by Lauritzen et al. [8] and has been used e.g. by Kyono [7].

**Theorem 4.4** (Minimal Covariate Adjustment). *Given a causal diagram $G = (V, E)$ and three pairwise disjoint vertex sets $X, Y, Z \subseteq V$, such that $G$ is $X$-loop-free, the following statements are equivalent:*

1. *$Z$ is a minimal covariate adjustment for identifying the causal effect from $X$ to $Y$.*
2. *$Z$ is a minimal set satisfying the adjustment criterion relative to $(X, Y)$.*
3. *$Z$ is a minimal set satisfying the back-door criterion relative to $(X, Y)$.*
4. *$Z$ is a minimal set satisfying the moral graph criterion relative to $(X, Y)$.*

As shown in Figure 2 the assumption that a DAG $G$ is additionally $X$-loop-free is essential.

We prove the theorem using the following auxiliary lemmata for an arbitrary causal diagram $G = (V, E)$ and pairwise disjoint sets of vertices $X, Y, Z \subseteq V$, such that $G$ is $X$-loop-free with respect to $X$. The considered criteria are relative to $(X, Y)$ in $G$.

**Lemma 4.5** ([12], Lemma 1). *If $Z$ satisfies the back-door criterion then $Z$ satisfies the adjustment criterion.*

**Lemma 4.6.** *If $Z$ is a minimal set satisfying the adjustment criterion then $Z$ contains no descendant of $X$.*

*Proof.* Assume that there exists $z \in Z$ which is a descendant in $G$ of a vertex in $X$. Note that from the adjustment criterion it follows that there is no directed path from $z$ to a vertex in $Y$. Moreover, from the assumption that $G$ is $X$-loop-free it follows that there is no directed path from $z$ to a vertex in $X$.

Let $W \subseteq V$ be the subset of all descendants of $z$ in $G$, including $z$. It is true that (1) there exists at least one node $z \in Z \cap W$, (2) in $G$ there exists no edge from a node in $W$ to a node in $V \setminus W$ (but there may be edges from nodes in $V \setminus W$ to nodes in $W$), and (3) $X \cup Y \subseteq V \setminus W$. The properties (1) and (2) follow easily from the definition of $W$. The fact that $Y \subseteq V \setminus W$ is true since otherwise there exists a causal path in $G$ connecting a node in $X$ with a node in $Y$ that is blocked. Finally $X \subseteq V \setminus W$ is true since $G$ is $X$-loop-free.

Now we remove all nodes $z \in Z \cap W$ from $Z$ and call the new set $Z'$. By the property (1) above $Z'$ is a proper subset of $Z$. We show that $Z'$ satisfies both statements (i) and (ii) of the adjustment criterion which contradicts the assumption that $Z$ is minimal.

Since $Z$ satisfies the statement (i) and $Z' \subseteq Z$ hence $Z'$ has to satisfy (i), too. Below we show that also statement (ii) is true, i.e. that any non-causal path $\pi$ in $G$ from $X$ to $Y$ is blocked by $Z'$.

To see this consider first that $\pi$ does not cross $W$, i.e. that $\pi$ does not contain any node in $W$. Since $\pi$ is blocked by $Z$ and it does not contain any node in $Z \setminus Z'$ (recall, we removed these nodes from $Z$) $\pi$ remains blocked by $Z'$. Next assume that $\pi$ reaches a node in $W$. The path has length at lest 2, i.e. $\pi$ consists of at least 3 vertices. Let $v_1, v_2, \ldots, v_k$, with $k \geq 3$, be the consecutive nodes along that path. Let $i$ be the smallest index such that $v_{i-1} \notin W$ and $v_i \in W$. From the property (3) above it follows that such an $i$ exists. Moreover, let $j > i$ be the smallest index such that $v_j \notin W$; again by (3) such an index exists, too. Now, from the property (2) we get that the edge incident to $v_{i-1}$ and $v_i$ has to be $v_{i-1} \to v_i$ and the edge incident to $v_{j-1}$ and $v_j$ has to be $v_{j-1} \leftarrow v_j$. Since neither $v_i, \ldots v_{j-1}$ nor their descendants belong to $Z'$, the path $\pi$ has to be blocked. $\square$

**Lemma 4.7.** *If $Z$ is a minimal set satisfying the adjustment criterion then $Z$ is a minimal set satisfying the back-door criterion.*

*Proof.* Assume $Z$ is a minimal set satisfying the adjustment criterion. By Lemma 4.6 we have that no element in $Z$ is a descendant of $X$. Moreover, by the property (ii) of the adjustment criterion it follows that $Z$ d-separates $X$ and $Y$ in $G_X$. Thus, $Z$ satisfies the back-door criterion. To prove the minimality, assume a proper subset $Z'$ of $Z$ satisfies the back-door criterion. From Lemma 4.5 it follows that $Z'$ satisfies the adjustment criterion – a contradiction to the assumption that $Z$ is a minimal set satisfying this criterion. $\square$

**Lemma 4.8.** *If $Z$ is a minimal set satisfying the back-door criterion then $Z$ is a minimal set satisfying the adjustment criterion.*

*Proof.* Assume $Z$ is a minimal set satisfying the back-door criterion. From Lemma 4.5 it follows that $Z$ satisfies the adjustment criterion. To see the minimality assume to the contrary that a minimal $Z' \subseteq Z$ satisfying this criterion is a proper subset of $Z$. By Lemma 4.7 we get that $Z'$ is a minimal set satisfying the back-door criterion – a contradiction. $\square$

*Proof (of Theorem 4.4).* We have proved that statements (1), (2) and (3) in Theorem 4.4 are equivalent; we now complete the proof by showing that statements (3) and (4) are equivalent. This means showing that a minimal adjustment $Z$ contains only variables from $An(X \cup Y) \setminus De(X)$. It was shown above that $Z$ contains no variables from $De(X)$; thus we are left with showing that $Z \subseteq An(X \cup Y)$. Suppose the converse, then there exists a $z \in Z$ such that $z \notin An(X) \cup An(Y) \cup De(X)$. Then removing all vertices $Z \setminus (An(X) \cup An(Y) \cup De(X))$ from $Z$ we get a proper subset $Z'$ of $Z$ and it is easy to see that $Z'$ still blocks all paths from $X$ to $Y$. Thus $Z$ is not minimal, a contradiction. $\square$

## 4.2 A Simplified Notion of $d$-Connectedness

In the following, we show that the rather intricate concept of $d$-connectedness can be translated to a simpler notion in a mixed graph. This will be useful later on to calculate the union of all biasing paths in an insufficiently adjusted causal diagram. Note that for this purpose, no special treatment of latent variables is necessary.

**Definition 4.9** (Fork). A path $v_1, \ldots, v_k$ in a mixed graph $G = (V, E)$, is called a *fork* if for some $i, j$ with $1 \leq i \leq j \leq k$, $v_i \to \ldots \to v_1$ and $v_j \to \ldots \to v_k$ are directed paths in $G$ and $v_i$—$\ldots$—$v_j$ is an undirected path in $G$.

For example, $t \to u \to v$ and $t \leftarrow u - v \to w$ are forks but $t \to u - v$ is not a fork.

**Definition 4.10** (Fork Graph). Given a DAG $G = (V, E)$ and three disjoint vertex sets $X, Y, Z \subseteq V$, the *fork graph* $G_\phi^Z = (V, E'), E' \subseteq \{(v, w), \{v, w\} \mid v, w \in V\}$ is constructed by performing for every edge $(v, w) \in E$ the following: (1) If $v \in Z$, remove $(v, w)$. (2) If $w \in An(Z)$, then substitute $(v, w)$ by an undirected edge $\{v, w\}$. (3) Otherwise leave $(v, w)$ unmodified.

E.g. for $G=$ 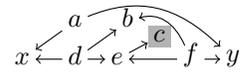 (where $Z = \{c\}$),

$G_\phi^Z=$ 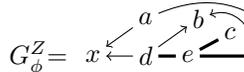 where the bold edges are those that were substituted by undirected edges.

**Theorem 4.11** (Mapping $d$-connected paths to forks). *Given a causal diagram $G = (V, E)$, three pairwise disjoint vertex sets $X, Y, Z \subseteq V$ and a path $\pi$ from $X$ to $Y$, then $\pi$ is d-connected in $G$ with respect to $Z$ if and only if $\pi$ is a fork in $G_\phi^Z$.*

*Proof.* First note that all paths from $X$ to $Y$ in $G_\phi^Z$ that contain no collider $u \to v \leftarrow w$ are forks. In fact, if there is no undirected edge on the path then we are done. Otherwise suppose $u \to v - w$ are violating edges; this would imply either that $v \to w$ is an edge in $E$ and $w$ is an ancestor of some $z \in Z$ or that $v \leftarrow w$ belongs to $E$ and $v$ is an ancestor of some $z \in Z$; in both cases we get the contradiction that $u \to v$ remains an edge in $G_\phi^Z$. If $u - v \leftarrow w$ are violating edges, we argue analogously.

Now let $\pi$ be any $d$-connected path in $G$. Note that $\pi$ remains a path in $G_\phi^Z$ since the only edges we remove from $G$ constructing $G_\phi^Z$ are the edges $(v,w)$ with $v \in Z$ and there will not be such an edge on a $d$-connected path from $X$ to $Y$. To show that $\pi$ is a fork in $G_\phi^Z$ we distinguish two cases. (1) $\pi$ contains no collider $u \to v \leftarrow w$ in $G$. Then $\pi$ is itself a fork in $G$ and constructing $G_\phi^Z$ any substitution of a directed edge by an undirected one implies that all its predecessor edges in a directed subpath become undirected thus yielding a fork in $G_\phi^Z$. (2) $\pi$ contains a collider $u \to v \leftarrow w$. Because $\pi$ is $d$-connected, the middle node $v$ is in $An(Z)$. Thus, in $G_\phi^Z$, the edges from $v$'s leftmost and rightmost ancestors in $\pi$ that point to $v$ are all undirected in $G_\phi^Z$. Iterating this argument yields that $\pi$ is a fork in $G_\phi^Z$.

For the other direction, let $\pi$ be any fork in $G_\phi^Z$. The undirected part of $\pi$ consists entirely of vertices from $An(Z)$, and no edge on $\pi$ points towards a vertex in $An(Z)$ on $\pi$ because it would otherwise have been deleted from $G_\phi^Z$. Hence $\pi$ is $d$-connected in $G$. □

## 5 Algorithmic Applications

The results of the previous section brought the criteria for covariate adjustment in causal diagrams closer to standard graph theory, which we can now exploit to obtain efficient algorithms for two important problems related to causal diagram analysis.

### 5.1 Enumerating Minimal Adjustment Sets

**Problem 5.1** (LIST-MINIMAL-ADJUSTMENTS).

**Input** A DAG $G = (V, E)$ and three pairwise disjoint vertex sets $X, Y, L \subseteq V$ such that $G$ is $X$-loop-free.

**Output** The set of all minimal covariate adjustments $Z \subseteq V$ that allow identification of the causal effect from $X$ to $Y$ and contain no variables from $L$.

Due to the equivalence of the adjustment and backdoor criteria in the minimal case (Theorem 4.4), we can solve the above problem by listing instead all minimal $d$-separators of $G_X$. Formally, this shows that the above problem can be reduced in linear time to the following one:

**Problem 5.2** (LIST-MINIMAL-D-SEPARATORS).

**Input** A DAG $G = (V, E)$ and three pairwise disjoint vertex sets $X, Y, L \subseteq V$ such that $G$ is $X$-loop-free.

**Output** The set of all minimal covariate sets $Z \subseteq V$ that $d$-separate $X$ from $Y$ in $G$ and contain no variables from $L$.

Algorithms for similar problems were presented by Acid and Campos [1] and Tian et al. [15]. These algorithms are either directly based on or very similar to the well-known Ford-Fulkerson-Algorithm for finding a separating set in an undirected graph. They are thus able to output *one* $d$-separator, if it exists, in worst-case polynomial time. Unfortunately, they do not lend themselves to generalization for outputting *all* $d$-separators.

In fact, a polynomial time algorithm that outputs all $d$-separators cannot exist for the simple reason that the number of these sets may be exponential in the size of the graph, such that the output alone would require more than polynomial time. This problem is addressed by so-called *polynomial delay* algorithms, whose complexity is measured not by their total running time, but per object output. Thus, if the number of solutions is polynomial, a polynomial delay algorithm will find and output them all in polynomial time. Otherwise, it can still produce a polynomially long list of different outputs in polynomial time, and the listing can be stopped and resumed at any time. This seems well suited to the problem at hand, because the adjustments are usually to be assessed by a *human* user, for whom a complete exponentially long list of options would be of little use. For more information on polynomial delay algorithms, we refer to the nice introduction in Takata's paper [13].

**Theorem 5.3** (Listing minimal $d$-separators with polynomial delay). *The problem* LIST-MINIMAL-D-SEPARATORS *can be solved by an algorithm that, after $O(n^3)$ preprocessing time, starts outputting the list of minimal $d$-separators using at most $O(n^3)$ processing time per element output, where $n$ is the number of variables in the input DAG.*

*Proof.* Applying the moral graph criterion, this boils down to a simple adaptation of an algorithm by Takata [13] that lists all minimal $X$-$Y$-separators of an *undirected graph* with polynomial delay $O(nm)$ and total space requirement $O(n)$, where $m$ is the number of edges in the graph. We only need to show that this algorithm can be extended to list only those minimal separators that do not contain any vertex from $L \cup W$ (where $W$ are the "forbidden vertices" from Theorem 4.2). For this purpose we can assume that $W \subseteq L$, i.e., we make all forbidden variables latent.

To ensure that forbidden vertices are not used in separators, we connect in the ancestor moral graph all variable pairs that are linked via forbidden vertices and then remove all forbidden vertices. Formally, let $G^m = (V, E)$ be the ancestor moral graph. Let the graph $G^L = (V \setminus L, E')$ be defined by $\{u, v\} \in E'$ iff (1) $\{u, v\} \in E$ or (2) there exists an undirected path $u$—$l_1$—...—$l_k$—$v$ in $G$ such that all intermediate vertices $l_i$ are in $L$. It is easy to see that the $X$-$Y$-separators of $G_L$ are precisely those $X$-$Y$-separators of $G^m$ that do not contain any vertex from $L$. $G^L$ can be constructed from $G$ in time $O(n^2)$.

Now, the following algorithm fulfills the properties claimed by the theorem: (1) Given the DAG $G$ and $X, Y, L$, construct the ancestor moral graph $G^m := (G_{\underline{X}}[\text{An}(X \cup Y)])^m$. (2) Construct $G^L$ from $G^m$ as explained above. (3) Apply Takata's algorithm to output all minimal $X$-$Y$-separators of $G^L$. The runtime of $O(n^3)$ is larger than the one of Takata's algorithm ($O(nm)$) because of the additional edges that are inserted into the moral graph and the $L$-transitive graph. The space requirement is $O(n)$, which is asymptotically optimal [13]. □

### 5.2 Identifying Bias in Insufficiently Adjusted Diagrams

Beyond minimal adjustments, causal diagrams provide an in-depth understanding of the "flow" of causal effects and bias via the causal and biasing paths they contain. The set of biasing paths, in particular, constitutes a *witness* of insufficient adjustment in a diagram and is thus useful for analyzing study design. To facilitate such analysis, the aforementioned programs *DAG program* and *dagR* contain algorithms that list all paths in a diagram. Such a list is easy to generate with polynomial delay using a standard backtracking approach. However, a full list of paths quickly becomes prohibitively long as the number of variables increases.

A strategy in such cases is to output a *compressed representation* of the path list rather than the list itself. The most natural compressed representation of a list of paths in a graph is probably the subgraph induced by these paths. Formally, we state our goal as solving the following problem:

**Problem 5.4** (IDENTIFY-BIASING-PATHS).

**Input** A DAG $G = (V, E)$ and three pairwise disjoint vertex sets $X, Y, Z \subseteq V$.
**Output** The list of all edges in $E$ lying on biasing paths from $X$ to $Y$ that are not blocked by $Z$.

This problem requires more effort to solve than one would perhaps expect. In particular, the following two approaches do *not* work.

First, one might consider constructing the ancestor moral graph $(G_{\underline{X}}[\text{An}(X \cup Y \cup Z)])^m$ and then labeling all undirected paths between $X$ and $Y$ in that graph. To see that this fails, consider the diagram in Figure 1a; in the ancestor moral graph, MR lies on an undirected path between LE and D even though it does not lie on a biasing path in the diagram (unless we adjust for MD like in Figure 1b). Second, one might consider using Shachter's "Bayes-Ball" algorithm [11] that performs an extended depth-first search to find out if $X$ and $Y$ are $d$-separated by $Z$. However, "Bayes-Ball" also would label edges that lie on non-simple biasing paths, i.e., biasing paths where edges can occur more than once. There appears to be no easy way of resolving these issues. However, the equivalence between $d$-connected paths and *forks* proved in Theorem 4.11 can be used to solve this problem in linear time.

**Theorem 5.5** (Identifying biasing paths in linear time). *The problem* IDENTIFY-BIASING-PATHS *can be solved in time* $O(|V| + |E|)$.

We will prove this theorem by showing that all $X$-$Y$-forks in the fork graph $(G_{\underline{X}})^Z_\phi$, obtained from $G$ by removing all edges emanating from $X$ and then applying Definition 4.10, can be identified in linear time. To this end, we generalize an algorithm presented by Eppstein [4, 5] to find disjoint directed paths from common ancestors to vertex sets $X$ and $Y$ in DAGs, which corresponds to the special case $Z = \emptyset$ of our problem. Eppstein's algorithm is based on computing for each vertex an index which we generalize to mixed graphs as follows:

**Definition 5.6** (Bottleneck number). Given the acyclic mixed graph $G = (V, E)$ and two disjoint vertex sets $X, Y \subseteq V$, let $T : V \to \mathbb{N}$ be a *topological numbering* of $G$, i.e. an index with $T(v) < T(w)$ if there is a directed path from $v$ to $w$ in $G$. Then the *bottleneck number* $B(v) \in \mathbb{N} \cup \{\bot\}$ is defined for every vertex $v$ as the largest index $T(w)$ of a vertex $w$ (possibly equal to $v$) through which all directed paths from $v$ to $X$ and all directed paths from $v$ to $Y$ go, if at least one such path exists. Otherwise $B(v) = \bot$.

E.g for $T(v) = 8 \overset{6}{\leftarrow} 2 \overset{5}{-} 3 \overset{4}{-\!\!-} 1 \to 7$, we get $B(v) = 8 \overset{6}{\leftarrow} 8 \overset{\bot}{-} \bot \overset{\bot}{-\!\!-} 7 \to 7$ (forks labeled bold) in the example introduced after Definition 4.10. Eppstein [4, Lemma 7] proved that for every $v \in V$, there are two disjoint directed paths from $v$ to $X$ and $Y$ if and only if $B(v) = T(v)$ (e.g. for the vertex with index 6 above). We generalize this lemma to acyclic mixed graphs through the following two lemmata. The proofs will appear in the full version of this paper.

**Lemma 5.7.** Let $G = (V, E)$ be an acyclic mixed graph and for every $v \in V$, let $B(v)$ denote the bottleneck number of $v$ with respect to $X, Y \subseteq V$. Then $v$ lies on a fork between $X$ and $Y$ if and only if one of the following three conditions holds: (1) $B(v) = T(v)$; (2) $v$ lies on an undirected path between two vertices $u, w$ with $B(u), B(v) \neq \bot$ and $B(u) \neq B(v)$; (3) $B(v) \neq \bot$ and $v$ is reachable via a directed path from another vertex $u$ for which one of the previous two conditions is satisfied.

**Lemma 5.8.** Let $G_\phi = (V, E)$ be the fork graph of some DAG $G$ and vertex set $Z$, and fix two disjoint vertex sets $X, Y \subseteq G$. Every edge $(u, v) \in E$ or $\{u, v\} \in E$ lies on a fork between $X$ and $Y$ if and only if both $u$ and $v$ lie on forks between $X$ and $Y$.

*Proof.* (Theorem 5.5) Given $G, X, Y, Z$, we first compute $(G_{\underline{X}})^Z_\phi$, which requires $O(|V| + |E|)$ time. Then we compute the bottleneck numbers $B(v)$ through a trivial extension of the $O(|V|+|E|)$ algorithm by Eppstein [4]. It remains to be shown that the vertices that lie on forks in $(G_{\underline{X}})^Z_\phi$ can be identified in linear time.

We first compute a list of all biconnected components with respect to the undirected edges in $(G_{\underline{X}})^Z_\phi$. For every such component $C$ with at least two vertices, we then perform the following on a copy of $C$: Find all vertices $v$ in $C$ with $B(v) \neq \bot$ and for every distinct value $n$ of $B(v)$, create two new vertices $s_n$ and $t_n$ with an edge $\{s_n, t_n\}$. For every $v$ with $B(v) \neq \bot$, create also an edge $\{t_{B(v)}, v\}$. Compute $C$'s block tree $C_{\text{block}}$, whose vertices correspond to the biconnected components and articulation points (vertices belonging to more than one biconnected component) in $C$. Every edge $\{s_n, t_n\}$ from $C$ corresponds to a leaf in $C_{\text{block}}$; label all these vertices, and afterwards label all vertices on paths between labeled vertices in the block tree. This corresponds to a labeling of the edges in $C$. Delete all edges that have not been labeled from $C$, and afterwards delete all isolated vertices except those with $B(v) = T(v)$. Every remaining vertex fulfills either condition (1) or condition (2) from Lemma 5.7.

Every vertex $v$ not belonging to one of the components treated above can be deleted from $(G_{\underline{X}})^Z_\phi$ if it does not fulfill either condition (1) or condition (3) from Lemma 5.7. A list of all vertices fulfilling condition (3) can be obtained by a single traversal of $(G_{\underline{X}})^Z_\phi$ after computing the bottleneck numbers. □

## 6 Conclusion

We studied several criteria for adjustment and $d$-separation in causal diagrams and obtained via these criteria fast algorithms for listing minimal adjustments and for identifying bias in causal diagrams. These algorithms form the basis for our online tool DAGitty (dagitty.net), which provides a graphical user interface for analyzing causal diagrams [14]. With the new algorithms, DAGitty is capable of analyzing causal diagrams with tens of variables, including some that were intractable by earlier programs, in real time. DAGitty's open source code can be consulted for additional reference on the algorithms presented here.

Future work could develop similar algorithms for other adjustment methods like front-door adjustment [9].

## References

[1] S. Acid and L. M. D. Campos. An algorithm for finding minimum d-separating sets in belief networks. In *Proceedings of UAI 1996*, pages 3–10, 1996.

[2] J. Berkson. Limitations of the application of fourfold tables to hospital data. *Biometrics Bulletin*, 2(3):47–53, 1946.

[3] L. Breitling. dagR: A suite of R functions for directed acyclic graphs. *Epidemiology*, 21(1):586–587, 2010.

[4] D. Eppstein. Finding common ancestors and disjoint paths in DAGs. Technical Report 95-52, Univ. of California, Irvine, 1995.

[5] D. Eppstein. Finding the k shortest paths. *SIAM J. Comput.*, 28(2):652–673, 1998.

[6] S. Knüppel and A. Stang. DAG program: identifying minimal sufficient adjustment sets. *Epidemiology*, 21(1):159, 2010.

[7] T. M. Kyono. Commentator: A front-end user-interface module for graphical and structural equation modeling. Technical Report R-364, University of California, Los Angeles, 1998.

[8] S. L. Lauritzen et al. Independence properties of directed markov fields. *Networks*, 20(5):491–505, 1990.

[9] J. Pearl. *Causality*. Cambridge U. Press, 2000.

[10] K. J. Rothman, S. Greenland, and T. L. Lash. *Modern Epidemiology*. Wolters Kluwer, 2008.

[11] R. D. Shachter. Bayes-ball: The rational pastime. In *Proceedings of UAI 1998*, pages 480–487, 1998.

[12] I. Shpitser, T. VanderWeele, and J. Robins. On the validity of covariate adjustment for estimating causal effects. In *Proceedings of UAI 2010*, pages 527–536. AUAI Press, 2010.

[13] K. Takata. Space-optimal, backtracking algorithms to list the minimal vertex separators of a graph. *Discrete Applied Mathematics*, 158:1660–1667, 2010.

[14] J. Textor, J. Hardt, and S. Knüppel. DAGitty: A graphical tool for analyzing causal diagrams. *Epidemiology*, 2011. In press.

[15] J. Tian, A. Paz, and J. Pearl. Finding minimal d-separators. Technical Report R-254, University of California, Los Angeles, 1998.